# Learning Non-Stationary Space-Time Models for Environmental Monitoring

**Sahil Garg** and **Amarjeet Singh**
Indraprastha Institute of Information Technology, Delhi

**Fabio Ramos**
School of Information Technologies, University of Sydney

## Abstract

One of the primary aspects of sustainable development involves accurate understanding and modeling of environmental phenomena. Many of these phenomena exhibit variations in both space and time and it is imperative to develop a deeper understanding of techniques that can model space-time dynamics accurately. In this paper we propose *NOSTILL-GP* - NOn-stationary Space TIme variable Latent Length scale GP, a generic non-stationary, spatio-temporal Gaussian Process (GP) model. We present several strategies, for efficient training of our model, necessary for real-world applicability. Extensive empirical validation is performed using three real-world environmental monitoring datasets, with diverse dynamics across space and time. Results from the experiments clearly demonstrate general applicability and effectiveness of our approach for applications in environmental monitoring.

## 1 Introduction

High fidelity understanding of environmental phenomena is crucial for sustainable development such that effective measures, including policy decisions, can be adapted accordingly. Such an understanding relies on models that can accurately represent dynamics in both space and time, as exhibited by many of environmental phenomena. Fig. 1 illustrates dynamics across both space and time for three real-world environmental phenomena - Ozone concentration, Wind speed and Indoor temperature. Several aspects of real environment affect the space-time dynamics thereby leading to a very large number of parameters (both known and unknown) when using a parametric modeling approach. Identification of these causal parameters and parameterizing a model with a large number of parameters are complex problems but can be addressed elegantly with Bayesian techniques. In particular, the nonparametric Bayesian framework is an excellent choice for modeling space-time dynamics in environmental phenomena due to its resilience to overfitting as complexity grows with new observations. We use Gaussian processes, GPs, (Rasmussen and Williams 2006) since they provide analytic forms for inference and learning that can be efficiently evaluated. GPs place a Gaussian prior over space of functions mapping inputs to outputs. As a Bayesian technique, GPs naturally balance data fit with model complexity while avoiding overfitting.

In this paper, we propose *NOSTILL-GP* - NOn-stationary Space TIme variable Latent Length scale GP to model natural phenomena. Our approach combines two critical aspects i.e. non-stationarity (varying dynamics at different locations) in both space and time and non-separability of space-time domain (accounting for dynamics in space affecting the dynamics across time and vice versa) to accurately model the environmental phenomena. The concept of variable latent length scale in our model intuitively extends a stationary model but brings in the complexity of learning a large number of model parameters that represent varying dynamics across space and time. A large number of parameters makes the learning process both complex as well as computationally intensive.

We propose several strategies for addressing the large number of parameters and prohibitive learning cost. These strategies include modeling the latent length parameters using a separate GP i.e. $GP_l$ (latent GP), intelligently selecting a small subset of locations for inducing latent GPs (using information gain and pseudo input concept), and a sparse representation of our GP model. We empirically validate the resulting models, after training them with these strategies, with three diverse datasets (as presented in Fig. 1), and demonstrate their scalability to large datasets (scalability in terms of smaller number of hyper-parameters). Empirical results clearly demonstrate effectiveness and general applicability of our technique in diverse environmental monitoring applications.

Specifically, the primary contributions of this work are: 1) A generic space-time GP model for accurately modeling environmental phenomenon; 2) Multiple strategies to address the prohibitive learning cost of our technique; 3) Extensive empirical validation using three diverse real-world environmental monitoring datasets.

## 2 Non-stationary space-time GP

A GP model places a multivariate Gaussian distribution over space of function variables, $f(\mathbf{x})$, mapping input space ($\mathbf{x} \in \Re^p$) to output space i.e. $f(\mathbf{x}) \sim \mathcal{GP}(m(\mathbf{x}), k(\mathbf{x}, \mathbf{x}'))$, where $m(\mathbf{x})$ specifies a mean function and $k(\mathbf{x}, \mathbf{x}')$ specifies a covariance function (also called kernel). As an example, for ozone data presented in Fig. 1a, the input space $\mathbf{x} \in \Re^3$ corresponds to a sensing location and a specific time instant for that sensing location; $m(\mathbf{x})$ represents the function that will provide mean ozone concentration at a given sensing location at a specific time instant; and $k(\mathbf{x}, \mathbf{x}')$ represents inter-dependency of ozone concentration at space-time locations $\mathbf{x}$ and $\mathbf{x}'$. The covariance function $k(\mathbf{x}, \mathbf{x}')$ is defined by a set of hyper-parameters $\Theta$ and written as $k(\mathbf{x}, \mathbf{x}'|\Theta)$ (Please refer to (Rasmussen and Williams 2006) for more details on covariance functions).

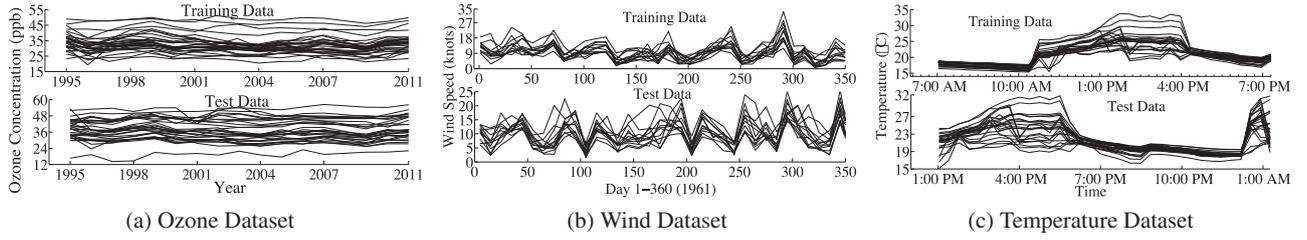

(a) Ozone Dataset  (b) Wind Dataset  (c) Temperature Dataset

Figure 1: Environmental monitoring datasets used as motivation and for validation. Each curve represents observations of the phenomenon at different times, at a specific location. We divide the locations into two sets, for training and testing purposes.

Learning a GP model is equivalent to determining the hyper-parameters of the covariance function from training data $\mathbf{y} \in \Re^n$ observed at $n$ locations $X \in \Re^{n \times p}$ (as an example, consider each of the curves in the training dataset in Fig. 1a as those representing $X \in \Re^{n \times 3}$). In a Bayesian framework, learning can be performed by maximizing log of marginal likelihood ($lml$), represented by

$$\log p(\mathbf{y}|X, \Theta) = -\frac{1}{2}\mathbf{y}^T K_y^{-1} \mathbf{y} - \frac{1}{2} \log |K_y| - \frac{n}{2} \log 2\pi,$$

where $K_y = K(X, X) + \sigma_n^2 I$; $\sigma_n^2$ representing observational noise. Conditioning on observed input locations $X$, the predictive distribution at unobserved locations $X_*$ can be obtained as $p(\mathbf{f}_*|X_*, X, \mathbf{y}) = \mathcal{N}(\boldsymbol{\mu}_*, \Sigma_*)$, where $\boldsymbol{\mu}_* = K(X_*, X)[K(X, X) + \sigma_n^2 I]^{-1}\mathbf{y}$, and $\Sigma_* = K(X_*, X_*) - K(X_*, X)[K(X, X)+\sigma_n^2 I]^{-1} K(X, X_*)$, where $K(X, X)$, $K(X_*, X_*)$ is prior on $X$, $X_*$ respectively; $\boldsymbol{\mu}_*$ and $\Sigma_*$ are predictive mean and posterior on $X_*$ conditioned on $X$ respectively. For example, we can use GPs to predict the distribution of curves presented as Test Data in Fig. 1a by training our model (i.e. learning the model hyper parameters, $\Theta$) using the curves presented as Training Data in Fig. 1a.

A typical assumption in GPs is that of stationarity i.e. the covariance function $K(\mathbf{x}, \mathbf{x}')$ only depends on distance between two locations $|\mathbf{x} - \mathbf{x}'|$. Moreover, to reduce complexity, when accounting for space-time variations, several techniques assume separability i.e. there is no space-time cross-correlation. However, real-world environmental applications are mostly non-stationary and would typically have space-time cross-correlations as well. Prior work (Singh et al. 2010) has shown that for space-time phenomena modeling, especially in environmental sensing applications, a special class of non-separable space-time covariance functions (Cressie and cheng Huang 1999; Gneiting 2002) is found to be more accurate than spatial (such as squared exponential, Matérn) or separable space-time covariance functions. Motivated by their evaluation and to realistically model the real-world environmental sensing applications, we focus on non-stationary, non-separable, space-time covariance functions for our *NOSTILL-GP* model.

Several techniques had been proposed in the literature for modeling non-stationary space-time phenomena (Rasmussen and Williams 2006; Ma 2003). While many of these techniques (e.g. periodic covariance function and nonlinear transformation of input space) perform well in specific environments, other non-stationary techniques (e.g. as proposed in (Ma 2003)) are generic but complex and less intuitive in the sense that it is difficult to analyze non-stationary phenomenon they may model well. Recently, several GP techniques had been proposed that model a phenomenon with variable correlation properties across input space thereby adapting to environment specific dynamics. This line of work includes techniques such as Mixture of GPs (Tresp 2001) and local kernel (Higdon, Swall, and Kern 1999; Paciorek and Schervish 2004; Plagemann, Kersting, and Burgard 2008). On similar lines, *NOSTILL-GP* employs the local kernel approach that does not make any assumptions about the dynamics of the phenomenon to be modeled and hence can adapt itself to accurately model the dynamics in an efficient manner.

## 3 *NOSTILL-GP* model

A non-stationary covariance function $K^{NS}(\mathbf{x}_i, \mathbf{x}_j) = \int_{\Re^2} K_{\mathbf{x}_i}(\mathbf{u}) K_{\mathbf{x}_j}(\mathbf{u}) \, d\mathbf{u}$, can be obtained by convolving spatially varying kernel functions, $K_{\mathbf{x}}(\mathbf{u})$ (Higdon, Swall, and Kern 1999) . Here $\mathbf{x}_i$, $\mathbf{x}_j$, and $\mathbf{u}$ are locations in $\Re^2$. For the specific case of Gaussian kernel functions, a non-stationary squared exponential covariance function $K^{NS}(\mathbf{x}_i, \mathbf{x}_j) = \sigma_f^2 |\Sigma_i|^{\frac{1}{4}} |\Sigma_j|^{\frac{1}{4}} \left|\frac{\Sigma_i + \Sigma_j}{2}\right|^{-\frac{1}{2}} \exp(-(\mathbf{x}_i - \mathbf{x}_j)^T \left(\frac{\Sigma_i + \Sigma_j}{2}\right)^{-1} (\mathbf{x}_i - \mathbf{x}_j))$ was derived (Paciorek and Schervish 2004), where the structure of the kernel remains the same and only the matrices $\Sigma_i$ and $\Sigma_j$ differ across input space. This extension was also generalized for other stationary covariance functions as:

$$K^{NS}(\mathbf{x}_i, \mathbf{x}_j) = |\Sigma_i|^{\frac{1}{4}} |\Sigma_j|^{\frac{1}{4}} \left|\frac{\Sigma_i + \Sigma_j}{2}\right|^{-1/2} K^S(\sqrt{q_{ij}}) \quad (1)$$

where $K^S(\tau)$ is a positive definite stationary covariance function on $\Re^p$ for every $p = 1, 2, ...$; $q_{ij} = (\mathbf{x}_i - \mathbf{x}_j)^T \left(\frac{\Sigma_i + \Sigma_j}{2}\right)^{-1} (\mathbf{x}_i - \mathbf{x}_j)$; $\Sigma_i$ and $\Sigma_j$ are local kernel ma-

289

trices at $\mathbf{x}_i$ and $\mathbf{x}_j$ input locations respectively and $\tau$ is the scaled distance between two input locations $\mathbf{x}_i, \mathbf{x}_j$.

Eq. 1 can be intuitively extended to a separable, stationary space-time covariance function by using time as another dimension. We, hereby, extend it to non-separable space-time covariance functions. Using $\tau = f(h, u)$, where $f$ is a deterministic function transforming scaled distance in space ($h$) and time ($u$) to scaled distance ($\tau$) in spatio-temporal domain, we can derive $\sqrt{q_{ij}} = f(\sqrt{q_{ij}^s}, \sqrt{q_{ij}^t})$ where $q_{ij}^s, q_{ij}^t$ are squared scaled distance in space and time respectively, with variable local kernel matrix across input space. Substituting expressions for $\tau$ and $q_{ij}$, we propose the following theorem:

**Theorem 1.** *Any stationary, non-separable, space-time covariance function $K_{ST}^S(h, u)$, when extended to $K_{ST}^{NS}$ as follows:*

$$K_{ST}^{NS}(\mathbf{x}_i, \mathbf{x}_j) = |\Sigma_i|^{\frac{1}{4}} |\Sigma_j|^{\frac{1}{4}} |(\Sigma_i + \Sigma_j)/2|^{-1/2}$$
$$\cdot K_{ST}^S(\sqrt{q_{ij}^s}, \sqrt{q_{ij}^t}) \quad (2)$$

*will remain a valid covariance function. Here $q_{ij}^s = (\mathbf{x}_{s_i} - \mathbf{x}_{s_j})^T \left(\frac{\Sigma_{i_s} + \Sigma_{j_s}}{2}\right)^{-1} (\mathbf{x}_{s_i} - \mathbf{x}_{s_j})$; $q_{ij}^t = (t_i - t_j)^2 \left(\frac{\Sigma_{i_t} + \Sigma_{j_t}}{2}\right)^{-1}$; $\mathbf{x} = [\mathbf{x}_s, t]$ (space and time coordinates); $\mathbf{x}_s \in \Re^p$ and $t \in \Re$; and $\Sigma_{i_s}, \Sigma_{i_t}$ are local kernel matrices at input location i for space and time respectively.*

*Proof Sketch.* This proof follows proof of Eq. 1 in (Paciorek and Schervish 2006). Class of functions positive definite on Hilbert space is identical with class of covariance functions of the form $K(\tau) = \int_0^\infty \exp(-\tau^2 r) dH(r)$ ($H(.)$ is non-decreasing bounded function; $r > 0$) (Schoenberg 1938). Using $q_{ij}$ as defined earlier, we get $K(\sqrt{q_{ij}}) = \int_0^\infty \exp(-q_{ij}r) dH(r) = \int_0^\infty \exp(-(\mathbf{x}_i - \mathbf{x}_j)^T \left(\frac{\Sigma_i}{r} + \frac{\Sigma_j}{r}\right)^{-1} (\mathbf{x}_i - \mathbf{x}_j)) dH(r)$. For the non-separable deterministic function $f(.)$, we get class of non-separable spatio-temporal covariance functions positive definite in hilbert space as $K(f(\sqrt{q_{ij}^s}, \sqrt{q_{ij}^t})) = \int_0^\infty \exp(-f((\mathbf{x}_{s_i} - \mathbf{x}_{s_j})^T \left(\frac{\Sigma_{i_s}}{r} + \frac{\Sigma_{j_s}}{r}\right)^{-1} (\mathbf{x}_{s_i} - \mathbf{x}_{s_j}), (t_i - t_j)^2 \left(\frac{\Sigma_{i_t}}{r} + \frac{\Sigma_{j_t}}{r}\right)^{-1})) dH(r)$. Expression $\exp(-f((\mathbf{x}_{s_i} - \mathbf{x}_{s_j})^T \left(\frac{\Sigma_{i_s}}{r} + \frac{\Sigma_{j_s}}{r}\right)^{-1} (\mathbf{x}_{s_i} - \mathbf{x}_{s_j}), (t_i - t_j)^2 \left(\frac{\Sigma_{i_t}}{r} + \frac{\Sigma_{j_t}}{r}\right)^{-1}))$ can be recognized as a non-stationary squared exponential covariance function which can be obtained from convolution of non-separable space-time Gaussian kernel $\frac{1}{(2\pi)^{\frac{p}{2}} |\Sigma_i|^{\frac{1}{2}}} \exp(-\frac{1}{2} f((\mathbf{x}_{s_i} - \mathbf{u}_s)^T \Sigma_{i_s}^{-1} (\mathbf{x}_{s_i} - \mathbf{u}_s), (t_i - u_t)^2 \Sigma_{i_t}^{-1}))$ (Paciorek and Schervish 2006) and we can define a stationary spatio-temporal covariance function as $K_{ST}^S(q_{ij}^s, q_{ij}^t) = K(f(\sqrt{q_{ij}^s}, \sqrt{q_{ij}^t}))$. Therefore, $K_{ST}^S(q_{ij^s}, q_{ij^t}) = \int_0^\infty \int_{\Re^p} k_{\mathbf{x}_i}(\mathbf{u}) k_{\mathbf{x}_j}(\mathbf{u}) d\mathbf{u} dH(r)$. Since $\int_{\Re^p} k_{\mathbf{x}_i}(\mathbf{u}) k_{\mathbf{x}_j}(\mathbf{u}) d\mathbf{u} > 0$ (Higdon, Swall, and Kern 1999), $K_{ST}^S$ and therefore $K_{ST}^{NS}$ is semi-positive definite (Paciorek and Schervish 2006). □

### 3.1 Latent length scale in *NOSTILL-GP*

An isotropic 1-D case for Eq. 1 was considered in (Plagemann, Kersting, and Burgard 2008) where $\Sigma_i$ simplifies to the latent length scale value $l_i^2$. Since real-world phenomena typically exhibit diverse variations across different dimensions, we consider 3-dimensional anisotropic case for our space-time generalization (Theorem 1) as:

$$\Sigma_i = \begin{bmatrix} \Sigma_{i_s} & 0 \\ 0 & \Sigma_{i_t} \end{bmatrix}, \Sigma_{i_s} = \begin{bmatrix} l_{x_i}^2 & 0 \\ 0 & l_{y_i}^2 \end{bmatrix}, \Sigma_{i_t} = l_{t_i}^2.$$

Here $l_{x_i}, l_{y_i}, l_{t_i} \in \Re$ are latent length scales at input location $\mathbf{x}_i$ in $x, y$, time dimensions respectively. Substituting expressions for local kernels in Eq. 2 and extending it for $X \in \Re^{n \times 3}$, we obtain

$$K_{ST}^{NS}(X, X) = P_r^{\frac{1}{4}} \circ P_c^{\frac{1}{4}} \circ \frac{P_s^{-\frac{1}{2}}}{8} \cdot K_{ST}^S(\sqrt{Q_s}, \sqrt{Q_t}), \quad (3)$$
$$P_r = P_{r_x} \circ P_{r_y} \circ P_{r_t}, \ P_c = P_{c_x} \circ P_{c_y} \circ P_{c_t},$$
$$P_s = (P_{r_x} + P_{c_x}) \circ (P_{r_y} + P_{c_y}) \circ (P_{r_t} + P_{c_t}),$$
$$P_{r_x} = \mathbf{l}_x^2 \cdot 1_n^T, \ P_{c_x} = 1_n \cdot \mathbf{l}_x^{2T}, \ Q_t = S_t^2 \circ ((P_{r_t} + P_{c_t})/2)^{-1},$$
$$Q_s = S_x^2 \circ ((P_{r_x} + P_{c_x})/2)^{-1} + S_y^2 \circ ((P_{r_y} + P_{c_y})/2)^{-1}.$$

Here, $P_{r_y}$ and $P_{r_t}$ are similar to $P_{r_x}$; $P_{c_y}$ and $P_{c_t}$ are similar to $P_{c_x}$; $\circ$ and $\div$ represent element wise matrix multiplication and division respectively; $\cdot$ represent matrix multiplication; $S_x, S_y, S_t \in \Re^{n \times n}$ (distances in $x, y,$ and time dimension respectively); and $\mathbf{l}_x, \mathbf{l}_y, \mathbf{l}_t \in \Re^n$ are latent length scale vectors for $X$. Note that Eq. 3 can be directly extended to any number of dimensions ($p$) in the space-time domain.

### 3.2 Strategies for efficient learning

The hyper-parameters, $\Theta$, for the non-stationary covariance function in Eq. 3 will be $\sigma_f, \mathbf{l}_x, \mathbf{l}_y, \mathbf{l}_t$. We make three observations here. Firstly, $\mathbf{l}_x, \mathbf{l}_y, \mathbf{l}_t$ are input space dependent. Secondly, using the learned values for $\mathbf{l}_x, \mathbf{l}_y, \mathbf{l}_t$ at training input space $X$, we need to infer $\mathbf{l}_{x_*}, \mathbf{l}_{y_*}, \mathbf{l}_{t_*}$ at test input space $X_*$. Finally, the number of parameters, in this case, is proportional to input size $n$ and hence learning the hyper-parameters will be computationally very expensive. We now propose several strategies for effectively learning our proposed *NOSTILL-GP* model.

**Using GPs to model latent length:** Motivated by a previously proposed approach to reduce the number of parameters in a GP model (Plagemann, Kersting, and Burgard 2008), we define three more independent GPs (and call them latent GPs) i.e. $GP_{l_x}, GP_{l_y}, GP_{l_t}$ for modeling the latent length scales in $x, y$ and time dimensions respectively. Each of these latent GPs can be modeled using simple covariance functions. We use sparse covariance functions (as explained later in detail) for latent GPs. $\mathbf{l}_x, \mathbf{l}_y, \mathbf{l}_t$ can then be calculated using the posterior distribution from the respective latent GPs (to ensure positive values for predicted latent length scales, latent GPs are modeled with log of latent length scale). To condition the latent GPs, we induce $m$ ($m << n$) input locations $\bar{X}$ as latent locations for which we assume that we know the latent length scale values. Combining the latent GPs and observation GP - $GP_y$, the latent length scale values at $m$ latent locations can be considered as parameters for the computationally efficient (4-GP) model representation (see Fig. 2) with a reduced number of parameters.

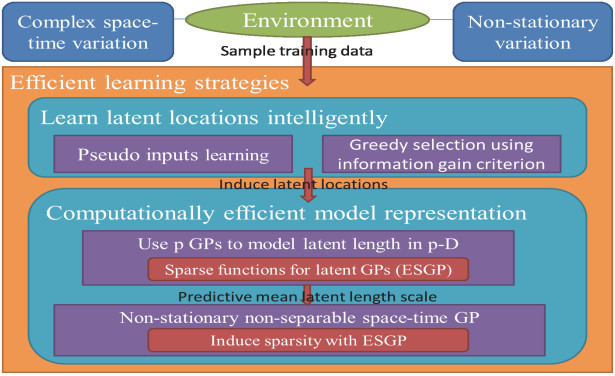

Figure 2: Illustration of efficient training of *NOSTILL-GP*

In this extended 4-GP model, we need to integrate over the predictive distribution of the latent length scale values to obtain predictive distribution for **y** in the input space $X$,

$$p(\mathbf{y}|X,\theta) = \int p(\mathbf{y}|X, \mathbf{l}_x, \mathbf{l}_y, \mathbf{l}_t, \theta_y) . p(\mathbf{l}_x|X, \bar{\mathbf{l}}_x, \bar{X}, \theta_{l_x})$$
$$. p(\mathbf{l}_y|X, \bar{\mathbf{l}}_y, \bar{X}, \theta_{l_y}) . p(\mathbf{l}_t|X, \bar{\mathbf{l}}_t, \bar{X}, \theta_{l_t}) \, dl_x \, dl_y \, dl_t,$$

where $\bar{\mathbf{l}}_x, \bar{\mathbf{l}}_y, \bar{\mathbf{l}}_t \in \Re^m$ are latent length scale vectors for $m$ induced input locations; $\theta_{l_x}, \theta_{l_y}, \theta_{l_t}, \theta_y$ are hyper-parameter sets for $GP_{l_x}, GP_{l_y}, GP_{l_t}, GP_y$ respectively. Since this integral is intractable, we follow the approximation of (Plagemann, Kersting, and Burgard 2008) considering only mean predictive values of the latent length scale variables. Thus, the predictive distribution for **y** is approximated to

$$p(\mathbf{y}|X,\theta) \approx p(\mathbf{y}|X, \mathbf{l}_x, \mathbf{l}_y, \mathbf{l}_t, \theta_y) \quad (4)$$

where $\mathbf{l}_x, \mathbf{l}_y, \mathbf{l}_t \in \Re^n$ are the predictive mean value vectors for the latent length scale in $x$, $y$ and time dimensions respectively, derived from $GP_{l_x}, GP_{l_y}, GP_{l_t}$ respectively.

Note that $\bar{\mathbf{l}}_x, \bar{\mathbf{l}}_y, \bar{\mathbf{l}}_t$ are hyper-parameters for $GP_y$ and observations for $GP_{l_x}, GP_{l_y}, GP_{l_t}$ respectively. (Plagemann, Kersting, and Burgard 2008) suggested learning parameters with an inner-outer loop approach (parameters for $GP_y$ and latent GPs are trained separably). The inner-outer loop approach gives sub-optimal solutions and might not converge. Instead, we learn the parameters for all the four GPs together. Further, since the latent length scales modeled by latent GPs are parameters for $GP_y$, as such there is no need to account for variance of the latent length scale predictions $\mathbf{l}_{x_*}, \mathbf{l}_{y_*}, \mathbf{l}_{t_*}$ at $X_*$ when predicting $\mathbf{f}_*$. Finally, since we use learned latent GPs to predict the latent length scales for the test input space, rather than directly mapping latent length scales from induced ($m$) locations, our proposed 4-GP model will not overfit in general.

**Intelligent latent location selection:** Since $m << n$, performance of *NOSTILL-GP* can be further optimized, if the latent locations are selected intelligently. We use two strategies for intelligently selecting latent locations - Information Gain and Pseudo Inputs.

*Using Information Gain for intelligent latent location selection:* We apply information gain concepts such as entropy and mutual information (MI) to greedily select $m$ latent locations that provide most information out of $n$ training locations. Greedy selection based on entropy and MI are shown to be near optimal under certain criterion (Krause, Singh, and Guestrin 2008). Entropy for a GP can be calculated in closed form. We greedily select the $(i+1)th$ latent location, after $X_i$ latent locations have been selected from training set containing $A$ location, that minimizes overall entropy of unselected locations, calculated as $H(X_{A/i+1}|(X_i \cup \mathbf{x}_{i+1})) \propto \log \|K(X_{A/i+1}, X_{A/i+1}|(X_i \cup \mathbf{x}_{i+1}))\|$ where $K(X_{A/i+1}, X_{A/i+1}|(X_i \cup \mathbf{x}_{i+1}))$ is the posterior covariance for the remaining unselected training locations, conditioned on the selected latent locations (Krause, Singh, and Guestrin 2008). Similarly, for the case of MI, we greedily select the $(i+1)th$ latent location that maximizes the mutual information, given by $I(X_{A/i+1}|(X_i \cup \mathbf{x}_{i+1})) = H(X_{A/i+1}) - H(X_{A/i+1}|(X_i \cup \mathbf{x}_{i+1}))$. For greedy selection of the most informative locations, we use following two covariance functions:

1. Empirical covariance: In this case, we first take data across all time steps to calculate empirical covariance matrix across spatial locations and greedily select $\bar{X}_s$ locations as latent locations across space. We then take the data across all spatial locations to calculate empirical covariance matrix across time and greedily select (separably) $\bar{X}_t$ latent locations across time. Finally, we combine $\bar{X}_s$, $\bar{X}_t$ to get induced latent locations across the space-time domain, $\bar{X}$.

2. Covariance function from a simple stationary space-time GP: can be used to greedily select (in an offline manner) the latent locations. Since the number of parameters in this case is small, the learning process is very fast. Using the learned stationary model, we first calculate covariance matrix across space-time locations and then greedily select latent locations, $\bar{X}$, non separably across space-time domain.

A clear benefit of using empirical covariance is that the *NOSTILL-GP* model does not rely on any stationary model for latent location selection. We also observed significantly reduced computational cost when using empirical covariance as compared to the stationary covariance model. However, a drawback of using empirical covariance is that latent locations can only be selected as a subset of training locations and can only be selected separably across space and time. Note that separable latent locations selection should not be confused with separable space-time GP modeling.

*Using Pseudo Input concept for intelligent latent location selection:* To reduce computation cost, (Snelson and Ghahramani 2006) parameterize the covariance function with data from induced $m$ pseudo inputs $\bar{X}$ ($m << n$). Hyper-parameters and the coordinates for $\bar{X}$ are learned by maximizing log of marginal likelihood ($lml$) where covariance function is $K_{nm}K_m^{-1}K_{mn} + diag(K_n - K_{nm}K_m^{-1}K_{mn})$ (Eqn. 9 of (Snelson and Ghahramani 2006)). Here $K_n$ is the prior calculated on $X$, and $K_m$ on $\bar{X}$. (Snelson and Ghahramani 2006) empirically showed that pseudo input locations $\bar{X}$ act as a better representative of training data than the greedy subset selection approach of (Seeger et al. 2003). They also demonstrated resilience of their approach to overfitting (even though number of parameters to be learned is large). We used their concept of pseudo inputs to intelligently learn latent locations $\bar{X}$. Any learned stationary space-time GP model can be used to calculate a prior on $X$ and $\bar{X}$. Based on the analysis given by (Snelson and Ghahramani 2006), we first learn the hyper-parameters and then learn coordinates for $\bar{X}$ from the training data keeping the hyper-parameters fixed.

**Exact sparse Gaussian process** To reduce the computational cost of GPs, several sparse approximation algorithms had been suggested in the past (Snelson and Ghahramani 2006; 2007; Quinonero-Candela and Rasmussen 2005). Considerable work in sparsity is based on the concept of induced input variables (Quinonero-Candela and Rasmussen 2005). (Melkumyan and Ramos 2009) introduced the concept of Exact Sparse GP (ESGP) by deriving an intrinsically sparse covariance function that compares well against other sparse approximations,

$$K^S(\tau) = \begin{cases} \sigma_f^2 \left[ \frac{2+\cos(2\pi\tau)}{3}(1-\tau) + \frac{1}{2\pi}sin(2\pi\tau) \right] & if \tau < 1 \\ 0 & if \ \tau \geq 1 \end{cases} \quad (5)$$

We achieve sparsity in our *NOSTILL-GP* model by performing element wise multiplication with the ESGP model from Eq. 5. ESGP assumes that a location is correlated to other locations only within a neighborhood. The size of the neighborhood is computed based on the length scale of the covariance function. For the sparse *NOSTILL-GP* model, the size of the neighborhood is variable across the input space because of variable latent length scale in this domain. We call this special property of the sparse *NOSTILL-GP* , Adaptive Local Sparsity. We also use Eq. 5 to model latent GPs for an efficient learning process. As a result, the computational cost for calculating the latent length predictive mean (which is $O(m^3)$ for a regular GP) is reduced significantly.

Fig. 2 illustrates the efficient learning strategies for the *NOSTILL-GP* model.

## 4 Experiments

We used three diverse environmental monitoring datasets (as shown in Fig. 1) to evaluate our proposed *NOSTILL-GP* model.

**USA Ozone Data:** Our first dataset is ozone concentration (in parts per billion) collected by United States Environmental Protection Agency[1] (Li, Zhang, and Piltner 2006). Due to several inconsistencies, we only selected data from year 1995 to 2011 (excluding data for 2007) for 60 stations across USA and used it for our evaluation purpose. For each station, we averaged ozone concentration for the whole year and took it as data for the corresponding station. We uniformly selected 30 out of 60 locations for training and remaining 30 locations for testing purposes. Fig. 1a shows the training and test data for Ozone concentration with each curve representing one station.

**Ireland Wind Data:** Our second dataset is daily average wind speed (in knots = 0.5418 m/s) data collected from year 1961 to 1978 at 12 meteorological stations in the Republic of Ireland[2] (Gneiting 2002). Our primary evaluation is on data from 1961 for all 12 stations with data from day 1 to day 351 (every 10 days) used for training and from day 5 to day 355 (every 10 days) used for testing purpose. Fig. 1b shows the training and test data for wind speed data of 1961 with each curve representing one station.

**Berkeley Intel Laboratory Temperature Data:** Our third dataset is from a deployment of 46 wireless temperature sensors in indoor laboratory region spanning 45 meters in length and 40 meters in width[3] at Intel Laboratory, Berkeley (Singh et al. 2010; Krause, Singh, and Guestrin 2008). Temperature data every 22 minutes, from 7 AM - 7:22 PM is used for training and from 1:00 PM to 1:22 AM (next day) for testing purpose. Out of 46 locations, we uniformly selected 23 locations each for training and testing purposes. Fig. 1c shows the training and test data with each curve representing temperature data collected from one of the sensors.

### 4.1 Model Selection

Based on prior work (Singh et al. 2010) demonstrating the effectiveness of a class of stationary, non-separable, space-time covariance functions from (Cressie and cheng Huang 1999; Gneiting 2002), we selected two such covariance functions (Ex. 1, 3 from (Cressie and cheng Huang 1999)) as presented by:

$$K_{ST}^S(h, u) = \frac{\sigma_f^2}{(u^2+1)^{\frac{p-1}{2}}} \exp\{-\frac{h^2}{u^2+1}\} \quad (6)$$

$$K_{ST}^S(h, u) = (\sigma_f^2 u^2 + 1)/\left[(u^2+1)^2 + h^2\right]^{p/2} \quad (7)$$

For the *NOSTILL-GP* model, we used Eq. 2 to convert Eq. 6, Eq. 7 into non-stationary, non-separable, space-time covariance functions. For modeling latent GPs in *NOSTILL-GP* model, we used covariance function from Eq. 5. We selected the approach from (Ma 2003)(we call it *NS-Chunsheng*), Mixture of GPs (Tresp 2001), Latent extension of input space (Pfingsten, Kuss, and Rasmussen 2006) to transform Eq. 6, Eq. 7 into a non-stationary, non-separable, space-time model, for comparative evaluation with our *NOSTILL-GP* model. To further reduce computation cost during inference, we induce sparsity to each of the covariance functions using the ESGP concept (Melkumyan and Ramos 2009).

### 4.2 Path Planning and Static Sensor Placement

*NOSTILL-GP* model can be used in both robotics and sensor networks to accurately represent space-time dynamics in an environment. Once a representative model for the environment is created; in robotics, optimal path planning for a mobile robot can be performed (Singh et al. 2010); while in sensor networks, optimal sensor placement at a subset of locations can be performed (Krause, Singh, and Guestrin 2008). To simulate applicability of our modeling approach for real-world setting, we selected the path planning problem to evaluate our modeling approach and compare it with other modeling approaches.

---
[1]http://epa.gov/castnet/javaweb/index.html
[2]http://lib.stat.cmu.edu/datasets/wind.desc
[3]db.csail.mit.edu/labdata/labdata.html

For path planning, we greedily (based on Entropy) select the next most informative location as per the model used. To closely simulate the real world setting, we only make observations at a few test locations during each time step (15, 6, 10 locations per timestep for ozone, wind and temperature data respectively). All the observations made in the past are then used to predict the phenomenon at unobserved locations, using the associated model, at any given time instant. Comparing the predicted value and ground truth value, we calculated the Root Mean Square (RMS) error and used it as the parameter to do comparative analysis of different modeling techniques.

### 4.3 Empirical Results

We performed several experiments, as shown in Table 1, considering different techniques for selecting latent locations.

| GP Model | Latent location selection approach | Referred to as |
|---|---|---|
| Stationary | NA | S |
| NS-Chunsheng | NA | NS-C |
| Mixture of GPs | NA | NS-MGP |
| Latent extension of input space | NA | NS-LEIS |
| NOSTILL | Greedy E | NS-GE |
| NOSTILL | Greedy MI | NS-GMI |
| NOSTILL | Greedy E (Emp) | NS-E-GE |
| NOSTILL | Greedy MI (Emp) | NS-E-GMI |
| NOSTILL | Pseudo Input | NS-P |
| NOSTILL | Uniform | NS-U |

Table 1: List of experiments: E - Entropy, MI - Mutual Information, Emp - Empirical covariance

| Experiment | Ozone | Wind | Temperature |
|---|---|---|---|
| S | 7.56 | 4.49 | 1.53 |
| NS-C | 7.27 | 3.83 | 2.42 |
| NS-M2GP | 4.01 | 4.50 | 21.75 |
| NS-M3GP | 9.00 | 4.34 | 10.46 |
| NS-LEIS | 4.01 | 4.46 | 1.75 |
| NS-E-GE | 2.90 | 2.56 | 1.68 |
| NS-E-GMI | 2.85 | 2.78 | 1.41 |
| NS-GE | 2.96 | 2.34 | 1.45 |
| NS-GMI | 2.96 | 2.73 | 1.47 |
| NS-P | 3.08 | 2.72 | 1.37 |
| NS-U | 4.00 | 2.72 | 1.54 |
| NS-GE-Eq. 6 | 2.82 | 2.74 | 1.36 |
| S-Eq. 6 | 7.39 | 3.74 | 2.92 |
| NSC-Eq. 6 | 7.50 | 3.76 | 2.45 |

Table 2: Mean RMS Error comparison for different models. M2GP, M3GP represents mixture of 2, 3 GPs respectively. Suffix Eq. 6 represents covariance function in Eq. 6. Rest of experiments are performed on covariance function in Eq. 7.

As mentioned in Sec. 3.2, we separably select space-time latent locations when using empirical covariance and non separably when using the stationary covariance function or when using the pseudo input concept. The number of latent locations selected ($m$) for Ozone data are 4-3 (separable), 12 (non separable), for wind data are 3-4 (separable), 12 (non separable) and for temperature data are 3-5 (separable) and 15 (non separable). Table 2 compares the mean of RMS error values calculated after each observation selection for different models and different datasets. Fig. 3 illustrates detailed comparison of *NOSTILL-GP* model (with empirical covariance used for latent location selection) with the corresponding stationary (S) model and another non stationary (NS-C) model ( number of latent locations $m$ (separable and non-separable) is specified as suffix in Fig. 3). We observe that *NOSTILL-GP* model performs consistently well for all three datasets while other approaches perform well only for some datasets. Within *NOSTILL-GP*, we observe that NS-GE performs more consistently than other approaches NS-U, NS-E-GE, NS-E-GMI, NS-P, NS-GMI, as shown in Table 2.

To analyze the effect of the number of latent locations, $m$, we performed experiments for NS-U, NS-P, NS-GE with varying $m$ for all three datasets (see Fig. 4a, 4b, 4c). We observe that *NOSTILL-GP* model starts performing well at a very small value of $m$ (optimum $m$ is approx. 12, 12 and 7 ($m \ll n$) for ozone, wind and temperature data respectively). Since temperature change is minimal across space-time during mornings and evenings (see Fig. 1c), a lower value for the optimum $m$ is expected. Fig. 4a and 4c also clearly indicate that uniform selection (NS-U) does not perform well when compared with either of the intelligent location selection techniques (NS-P, NS-GE).

We observe that the performance of NS-P is similar to NS-GE, even though the computational cost for learning latent locations for NS-P is higher. Therefore we recommend using the simple greedy approach for selecting the latent locations unless $m$ is very small.

To further test the general applicability of the *NOSTILL-GP* model across different test input space, we trained different models (S, NS-C and NS-E-GE-3-4) with wind data of year 1961, and tested with data of years 1961, 1963, 1966, 1969, 1972, 1975 and 1978 (36 days selected uniformly as timesteps in each year and all of 12 stations selected across space). Fig. 4d compares the performance of the three models across data from different years. We observe that *NOSTILL-GP* model performs consistently better than other models across all the years (note that latent locations were learned in training input space from training data only).

## 5 Conclusion

We proposed a generic approach for creating non-stationary non-separable space-time GP model, *NOSTILL-GP*, for environmental monitoring applications. We further proposed different strategies for efficiently training of our model thus making it applicable for real-world applications. We performed extensive empirical evaluation using diverse and large environmental monitoring datasets. Experimental results clearly indicate the scalability, consistency and general applicability of our approach for diverse environmental monitoring applications.

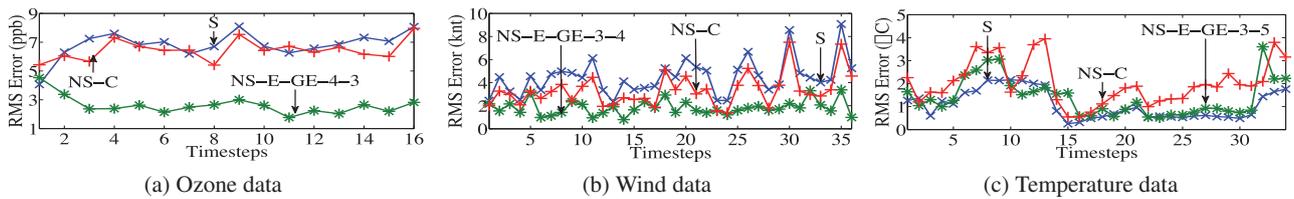

Figure 3: Root Mean Square (RMS) error comparison for different modeling approaches for datasets as shown in Fig. 1

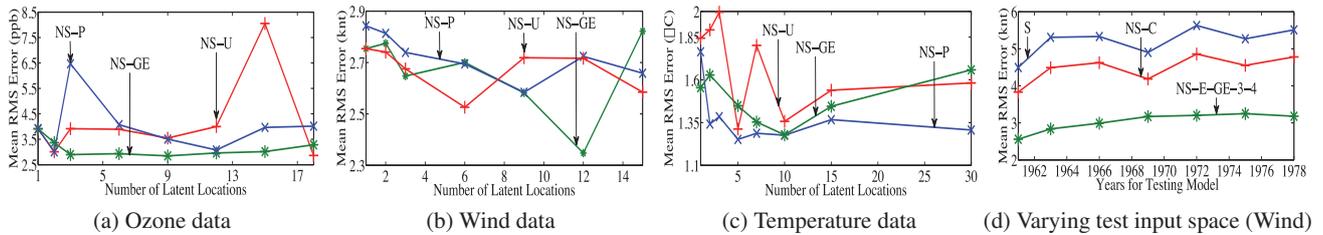

Figure 4: Mean RMS variation with change in number of latent locations ($m$) and change of test input space

## 6 Acknowledgments

Sahil Garg and Amarjeet Singh were partially supported through a grant from Microsoft Research, India. Amarjeet Singh was also supported through IBM Faculty Award.